\pdfoutput=1

\documentclass[11pt]{article}

\usepackage{acl}

\usepackage{times}
\usepackage{latexsym}

\usepackage[T1]{fontenc}

\usepackage[utf8]{inputenc}

\usepackage{microtype}
\usepackage{amsmath}
\usepackage{amssymb}
\usepackage{inconsolata}
\usepackage{color}
\usepackage{xcolor}
\usepackage{colortbl,booktabs}
\usepackage{multicol}
\usepackage{multirow}
\usepackage{booktabs}
\usepackage{pifont}
\usepackage{wrapfig}
\usepackage{graphicx} 
\usepackage{comment}
\usepackage{color}
\usepackage{xcolor}
\usepackage{algorithmic}
\usepackage{enumitem}
\usepackage[ruled, linesnumbered]{algorithm2e}

\title{DSEE: Dually Sparsity-embedded Efficient Tuning of Pre-trained Language Models}

\author{%
  Xuxi Chen\textsuperscript{1}, Tianlong Chen\textsuperscript{1}, Weizhu Chen\textsuperscript{2}, \textbf{Ahmed Hassan Awadallah}\textsuperscript{2}\\
  \textbf{Zhangyang Wang}\textsuperscript{1}, \textbf{Yu Cheng}\textsuperscript{2}\\
  {\textsuperscript{1}University of Texas at Austin, \textsuperscript{2}Microsoft Corporation}\\
  \texttt{\{xxchen,tianlong.chen,atlaswang\}@utexas.edu} \\ \texttt{\{wzchen,hassanam,yu.cheng\}@microsoft.com} \\
}

\begin{document}
\maketitle
\begin{abstract}
Gigantic pre-trained models have become central to natural language processing (NLP), serving as the starting point for fine-tuning towards a range of downstream tasks. However, two pain points persist for this paradigm: (a) as the pre-trained models grow bigger (e.g., $175$B parameters for GPT-3), even the fine-tuning process can be time-consuming and computationally expensive; (b) the fine-tuned model has the same size as its starting point by default, which is neither sensible due to its more specialized functionality, nor practical since many fine-tuned models will be deployed in resource-constrained environments. To address these pain points, we propose a framework for resource- and parameter-efficient fine-tuning by leveraging the sparsity prior in both weight updates and the final model weights. Our proposed framework, dubbed \textbf{D}ually \textbf{S}parsity-\textbf{E}mbedded \textbf{E}fficient Tuning (DSEE), aims to achieve two key objectives: (i) \textit{parameter efficient fine-tuning} -  by enforcing sparsity-aware low-rank updates on top of the pre-trained weights; and (ii) \textit{resource-efficient inference} - by encouraging a sparse weight structure towards the final fine-tuned model. We leverage sparsity in these two directions by exploiting both unstructured and structured sparse patterns in pre-trained language models via
a unified approach. 
Extensive experiments and in-depth investigations, with diverse network backbones (i.e., BERT, RoBERTa, and GPT-2) on dozens of datasets, consistently demonstrate impressive parameter-/inference-efficiency, while maintaining competitive downstream performance. For instance, DSEE saves about $25\%$ inference FLOPs while achieving comparable performance, with $0.5\%$ trainable parameters on BERT. Codes are available at \url{https://github.com/VITA-Group/DSEE}.
 
\end{abstract}

\section{Introduction}
Most recent NLP applications have been following the pre-train then fine-tune paradigm, starting from a gigantic pre-trained model and fine-tuning it towards downstream tasks. Conventional \textit{fine-tuning} works through updating all of the parameters in the pre-trained model. However, as the size of pre-trained models grows, updating all parameters becomes less feasible in most practical scenarios, due to the expensive memory and computational requirements. For example, BERT$_\mathrm{BASE}$~\citep{devlin2019bert} has $110$M trainable parameters, while GPT-2~\citep{radford2019language} has up to $1.5$B and the largest version of GPT-3~\citep{radford2019language} has an astonishing $175$B trainable parameters. As such, conventional fine-tuning of the larger models could require hundreds of GPU hours. Another downside of this paradigm is that it requires storing as many parameters as in the large-scale pre-trained models for each downstream task, which poses impediments to the deployment in real-world resource-constrained environments. 


One solution to address the extensive resource requirement of conventional fine-tuning is model pruning~\citep{lecun1990optimal,han2015deep,ren2018admmnn,he2017channel,liu2017learning}, where unnecessary weights are eliminated to shrink the model size. For example, ~\citet{chen2021earlybert} leverages $\ell_1$ regularization to remove insignificant attention heads and gains $35\sim45\%$ training time with comparable performance;  \citet{chen2021pixelated,dao2022monarch} leverage sparse matrices with fixed structures to reduce pretrained models' sizes. All these studies indicate the rise of sparsity naturally during fine-tuning a general-purpose pre-trained model, to some specialized downstream functionality. One potential interpretation, of why sparsity arises, is that different subsets of the parameters may be responsible for different downstream tasks and data domains~\citep{sanh2020movement}. However, identifying appropriate sparse masks can be burdensome: fine-tuning a large pre-trained language model like GPT-3 for just one step consumes at least $1.2$TB of VRAM and requires $96$ pieces of NVIDIA Tesla~\citep{hu2021lora}, and these methods either require access to pre-trained weights or introduce additional learnable coefficients (such as importance scores of attention heads). 



A parallel alternative is to design \textit{parameter-efficient} fine-tuning algorithms, which aim at optimizing a small portion of weights while fixing most of them when fine-tuning on downstream tasks. Pioneering works along this line, which utilize adapters~\citep{houlsby2019parameter}, learnable embeddings~\cite{li2021prefix,liu2021p}, low-rank decomposition~\citep{hu2021lora} or their combination~\cite{he2021towards}, can significantly reduce the number of trainable parameters while preserving good fine-tuning performance. Although these methods can substantially improve the storage and deployment efficiency of models, there are two major hurdles: ($i$) they does not yield any inference efficiency gains since the full pre-trained weights are still required to calculate outputs; and ($ii$) current methods assume the updates on pretrained weights to be either sparse~\citep{guo2020parameter} or low-rank~\citep{hu2021lora}, yet those assumptions might be oversimplified~\citep{yu2017compressing} and overly restricted to allow for effective updates.
These observations have inspired us to explore better parameter-efficiency methods.

To improve both resource- and parameter-efficiency during model fine-tuning, we explicitly draw on the prior of sparsity for \textit{both weight updates and the final weights}, and establish a \textit{dually sparsity-embedding efficient tuning} (\textbf{DSEE}) framework. Starting from a pre-trained model, DSEE first adopts a sparsity-aware low-rank weight update to achieve \textit{parameter-efficiency} of the fine-tuning process; and then enforces a sparse weight structure directly from weight updates by masking to achieve \textit{resource-efficiency} of the fine-tuned model at inference time. Our contributions can be summarized as follows:


\vspace{-2mm}
\begin{itemize}[leftmargin=*]

    \item We propose the dually sparsity-embedding efficient tuning, which unifies sparsity-aware parameter-efficient weight update and sparse pretrained weight in fine-tuning gigantic pre-trained models. It is the first attempt toward jointly optimizing both parameter-efficiency of the fine-tuning process and the resource-efficiency of the fine-tuned model.
    \item 
    
    Both unstructured and structured sparse priors are investigated in our proposed DSEE algorithm.  \underline{For weight updates}, the injected sparsity prior enhances existing parameter-efficient update schemes (e.g., low-rank decomposition). \underline{As for the final weights}, we draw superior sparse masks, either unstructured or structured, directly from the weight updates, which requires neither additional parameters nor access to the pre-trained weights and saves the sparsification cost. 
    \item Extensive experiments demonstrate the effectiveness of our proposal across various representative pre-trained language models (BERT, GPT-2, and RoBERTa) and on diverse evaluation benchmarks (E2E, DART, WebNLG, and GLUE). On GPT-2, our methods can achieve a BLUE score of $\{$69.5$, $54.9$, $47.5$\}$ with $0.1\%$ of trainable parameters on \{E2E, WebNLG, DART\} with $20\%$ parameter removed in pre-trained weights. On BERT, DSEE can fine-tune only $0.5\%$ parameters and save about $25\%$ inference FLOPs, while losing less than $2\%$ performance. 
\end{itemize}

\vspace{-2mm}
\section{Related Work}
\vspace{-1mm}

\paragraph{Pruning and Sparsification} 
Pruning is a classical model compression technique that can reduce the number of parameters, which can bring training and inference efficiency. Researchers have proposed several pruning methods for pre-trained language models: \citet{mccarley2019structured, chen2021earlybert} pruned attention heads that had less contribution during finetuning; \citet{sanh2020movement} proposed a pruning criterion targeting the weight change after training, which suits the transfer learning better; \citet{wang2020structured} incorporated low-rank factorization and $\ell_0$ regularization for pruning. Recently, there is a series of sparsification works that utilize sparse masks with specific structures, called Butterflies, and achieve high efficiency in pretraining models~\cite{chen2021pixelated} or fine-tuning on downstream tasks~\cite{dao2022monarch}. However, these methods do not allow for parameter-efficient updates. 

\paragraph{Low-rank decomposition}
Low-rank approximation~\citep{ye2005generalized} has broad applications in the machine learning community and is vastly studied. One classical scenario is the robust principal component analysis~\citep{candes2011robust}, which decomposes a matrix into a low-rank plus a sparse component. 
Existing literature shows that in deep learning, the learned over-parameterized models often naturally bear approximate low-rank weight structures~\citep{oymak2019generalization,yu2017compressing}. Some~\citep{jaderberg2014speeding,povey2018semi,sainath2013low,zhang2014extracting,zhao2016low} have explicitly imposed the low-rank constraint during training. \citet{wang2020structured,hu2021lora} utilized low-rank decomposition to shrink the model size and trim down the trainable parameters during fine-tuning. However, to our best knowledge, integrating sparsity and low-rank structures has never been studied before for efficient fine-tuning of pre-trained language models. 






\paragraph{Parameter-efficient adaptation.} 

Parameter-efficient fine-tuning aims to reduce the number of trainable parameters when fine-tuning the models across different downstream domains. Unlike pruning, it aims at adapting models with fewer parameters instead of building sparse models. Various approaches are proposed to achieve this goal:
\citet{rebuffi2017learning,houlsby2019parameter} inserted and only trained adapters between existing layers, whose parameters are much less compared to the pretrained models. \citet{guo2020parameter} leveraged $\ell_0$ regularization to limit the number of non-zero elements in the update vectors. \citet{lester2021power,li2021prefix,liu2021p} introduced efficient prompt tuning which optimizes only a small continuous task-specific vector. \citet{zaken2021bitfit} fine-tunes only the bias terms inside models. \citet{hu2021lora} proposed a low-rank decomposition-based method, and \citet{he2021towards} combined low-rank and adapter-based methods for efficient finetuning.  However, fine-tuned models yielded by these methods have the same amount of weights as the pre-trained models; hence they contribute no resource efficiency. 
\vspace{-2mm}
\section{Methodology}
\vspace{-1mm}
In this section, we describe our notations and definitions of sparsity generation and parameter-efficient fine-tuning in Section~\ref{sec:notations}, and then introduce the dually sparsity-embedded efficient fine-tuning algorithms in Sections~\ref{sec:sparsity_plus} and~\ref{sec:DSEE}. 

\begin{figure}[t]
\centering
\vspace{-1.5em}
\includegraphics[width=1\linewidth]{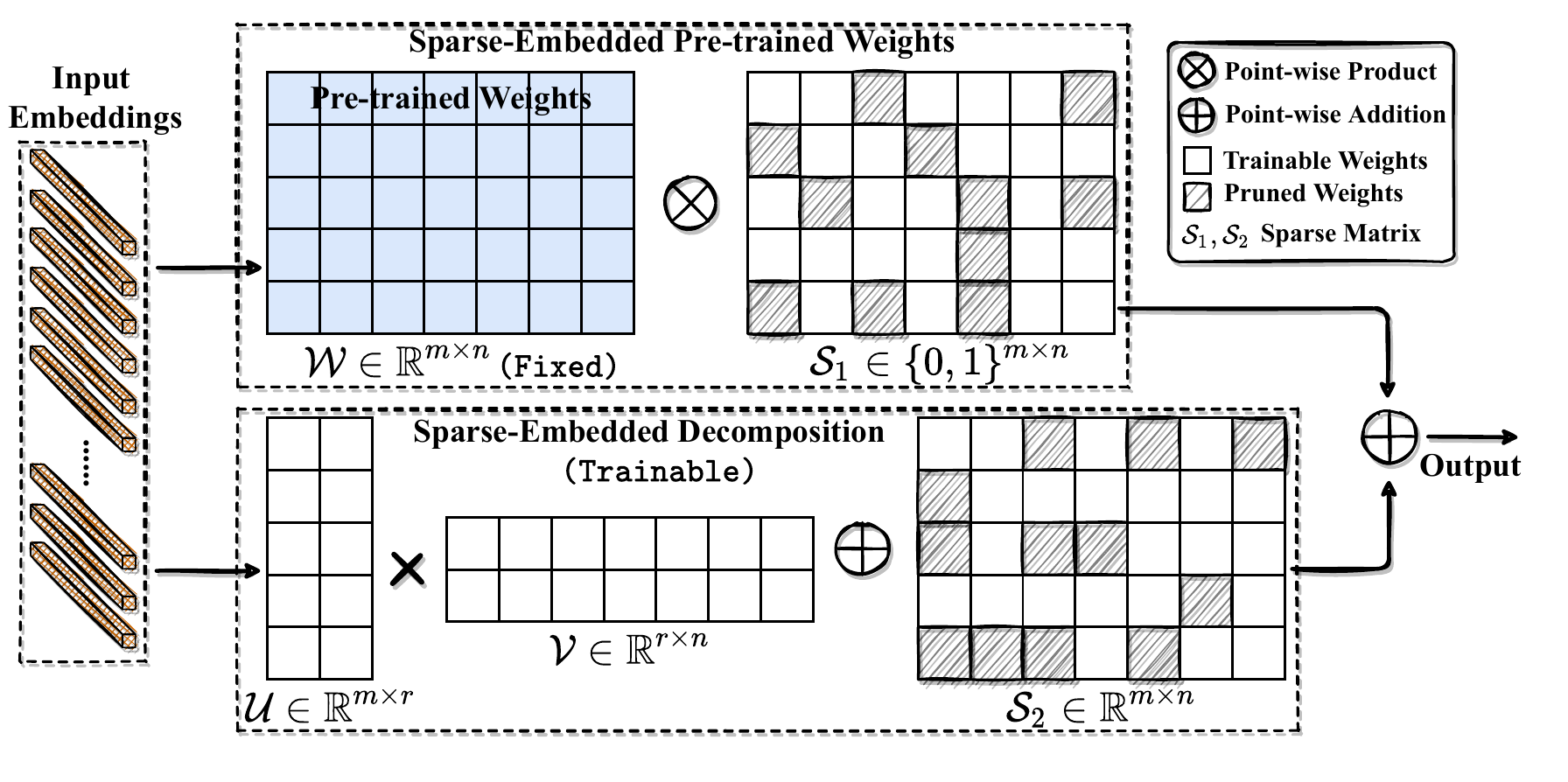}
\vspace{-9mm}
\caption{{\small The overview of our proposal. The sparse masks can have unstructured or structured patterns, which leads to resources efficiency. During the fine-tuning, we only train decomposed matrices $\mathcal{U}$, $\mathcal{V}$ and non-zero elements in $\mathcal{S}_2$.}}
\vspace{-4mm}
\label{fig:methods}
\end{figure}

\subsection{Preliminaries} \label{sec:notations}

\paragraph{Sparsification and resource-efficient fine-tuning.} We adopt both unstructured and structured pruning methods to produce sparsity. They can lead to resource-efficiency including memory and computation savings.  
 
Given $\mathcal{W}\in\mathbb{R}^{m\times n}$ a weight matrix, pruning aims at finding a binary mask $\mathcal{M}\in\{0,1\}^{m\times n}$ which is applied to $\mathcal{W}$ and generating a sparse weight $\mathcal{W}\odot\mathcal{M}$. The weights at the positions where $\mathcal{M}$ have ``0'' value are considered as pruned. Pruning methods can be classified into two classes by the structure of $\mathcal{M}$: For unstructured pruning where $\mathcal{M}$ does not have sparse structures such as rows and columns, the memory cost is saved due to fewer number of nonzero parameters; for structured pruning, it also helps save computational cost since the sparse weights can be smaller in size. One of the most widely used unstructured pruning methods is the weight magnitude~\cite{han2015deep}, \textit{i.e.}, remove the weights with the smallest absolute values. One common structured pruning method in the NLP field is the head pruning~\cite{mccarley2019structured}, which tries to remove unimportant attention heads from the model.  

\vspace{-3mm}
\paragraph{Parameter-efficient fine-tuning.} To leverage the knowledge in pre-trained weights $\mathcal{W}$, downstream models learn task-specific weight update $\Delta\mathcal{W}$ via fine-tuning and generate predictions with weights $\mathcal{W}+\Delta \mathcal{W}$, where the output of models is calculated as $(\mathcal{W} + \Delta \mathcal{W})x$ with $x$ as the input. Since $\Delta\mathcal{W}$ has the same size as $\mathcal{W}$, learning the update matrices usually requires massive resources as the size of the pre-trained model increases. Parameter-efficient fine-tuning tries to solve this problem by using as few trainable parameters as possible to represent $\Delta\mathcal{W}$, while maintaining competitive downstream fine-tuning performance. Previous literature reaches the goal via either sparsifying weight update matrices $\Delta\mathcal{W}$~\citep{guo2020parameter} or leveraging low-rank decomposed matrices to compute $\Delta\mathcal{W}$~\citep{hu2021lora}, while in our work we combine both of them. 

\begin{algorithm} 
\SetAlgoLined
\SetNoFillComment
\KwIn{Pretrained weights $\mathcal{W}$, number of non-zero elements $N$, number of weights to decompose $n$. }
\KwOut{Indices sets $\Omega_i,i=1,2,\dots,n$. }
Initialize each $\Omega_i$ to be an empty set.  \\
\For{each weight matrix $\mathcal{W}_i$ in $\mathcal{W}$}{
\tcc{{\small Decomposition}}
Perform matrix decomposition $\mathcal{W}_i\approx \mathcal{A}\mathcal{B}+\mathcal{S}'$ by solving the optimization problem~\ref{eqn:decompose}. \\
\tcc{{\small Extract important elements from $\mathcal{S}'$ into $\Omega_i$. }}
Perform thresholding on $\mathcal{S}'$: Keep $N$ elements in $\mathcal{S}'$ with top magnitudes, and append their locations into $\Omega_i$.  \\
}
\caption{Sparsity-Embedded Low-Rank Decomposition}
\label{alg:sparse_lora}
\end{algorithm}
\vspace{-3mm}


\subsection{Sparsity-Embedded Parameter-Efficient Fine-tuning} \label{sec:sparsity_plus}


A recent study~\citep{hu2021lora} enforces low-rank constraint to weight update tensors $\Delta\mathcal{W}$, and obtains a satisfactory trade-off between parameter-efficiency and model quality. However, as revealed experimentally by~\citep{yu2017compressing}, a part of the important information in the trained weights scatters outside the low-rank subspace, creating sparse ``residuals". Inspired by this observation, we investigate a new sparsity-aware low-rank subspace of $\Delta\mathcal{W}$, and introduce the first component of our proposal in Figure~\ref{fig:methods}, \textit{i.e.}, sparsity-embedded parameter-efficient fine-tuning.

Specifically, the weight updates $\Delta\mathcal{W}$ are consisted of two components as illustrated in Figure~\ref{fig:methods},: (1) a low-rank component $\Delta\mathcal{W}_l$ built by the multiplication of two matrices $\mathcal{U}\in\mathbb{R}^{m\times r}$ and $\mathcal{V}\in\mathbb{R}^{r\times n}$; and (2) a sparse residual $\Delta\mathcal{W}_s=\mathcal{P}_\Omega (\mathcal{S})$ where $\mathcal{S}\in\mathbb{R}^{m\times n}$ is a learnable matrix, $\mathcal{P}_{\Omega}(\mathcal{S}) = 
\left\{
\begin{aligned}
s_{i,j}&, \quad(i,j)\in \Omega \\
 0&,  \quad(i,j)\in \Omega^{C}
\end{aligned}
\right.,\ i=1,2,\dots,m,\ j=1,2,\dots,n$, $w_{i,j}$ is the parameter of $\mathcal{S}$ at location $(i,j)$, and $\Omega$ is a indices set containing the positions of non-zero elements in $\mathcal{S}$.  
 The update matrix $\Delta\mathcal{W}$ is expressed as $\Delta\mathcal{W}_l+\Delta\mathcal{W}_s$, with $\mathcal{U}$, $\mathcal{V}$ and $\mathcal{S}$ as the learnable parameters while $\Omega$ is fixed once determined. Compared to the full fine-tuning which has $m\times n$ trainable parameters for a matrix with size $m\times n$, our method only has $(m+n)\times r + \mathrm{card}(\Omega)$ trainable parameters. If $r$ is smaller than $\frac{m\times n-\mathrm{card}(\Omega)}{m + n} \lessapprox 0.5 \min\{m,n\}$, our method is capable of reducing trainable parameters for downstream fine-tuning. In practice, the value of $r$ is very small compared to $m$ and $n$ so the savings are significant.  

One question for the above pipeline is how to find a high-quality indices set $\Omega$. Inspired by the observation that the low-rank component $\Delta\mathcal{W}_l$ is highly correlated with the low-rank structure of $\mathcal{W}$~\cite{hu2021lora}, we hypothesize that the indices set $\Omega$ should be highly correlated as well. More concretely, we hypothesize that the sparse residuals that are not in the low-dimensional subspace of $\mathcal{W}$ may also lay outside $\Delta\mathcal{W}_l$, which motivates the design of sparse update $\Delta\mathcal{W}_s$. We formulate the problem of discovering the sparse residuals of $\mathcal{W}$ as a Robust Principal Component Analysis~\cite{candes2011robust}. Formally, we aim at solving the following optimization problem: 
\begin{equation}
    \label{eqn:decompose}
    \begin{aligned}
    \min_{U,V,S} \;\; & \frac{1}{2} \| W - UV - S\|_F^2 \\
    \textit{s.t.} \;\; & \mathrm{rank}(U) \leq r,\ \mathrm{rank}(V) \leq r, \\& \mathrm{card}(S) \leq c. 
    \end{aligned}
\end{equation}
where $\mathrm{rank}(\cdot)$ and $\mathrm{card}(\cdot)$ indicate the rank and the number of non-zero elements of a matrix, respectively.
$\mathcal{S}'$ represents the sparse residuals that cannot be fit in the low-rank component $\mathcal{A}\mathcal{B}$, and we acquire the locations of elements with non-zero magnitude into $\Omega$. 
To solve Problem~\ref{eqn:decompose} efficiently, we adopt an SVD-free algorithm called GreBsmo~\citep{zhou2013greedy} (refer to Section~\ref{sec:decompose}). 
Algorithm~\ref{alg:sparse_lora} summarizes the detailed procedure of constructing sparse indices sets $\Omega$.  
Empirically, we set the size of $\Omega$ (\textit{i.e.}, $c$) to be $16$ since it yields high test performance (refer to Section~\ref{sec:abl}) while imposing little overhead on parameters. The initial values of $\mathcal{V}$ and $\mathcal{S}$ are set as $0$ so these matrices do not affect outputs at the beginning of training.

\subsection{Dually Sparsity-Embedded Efficient Tuning (DSEE)} \label{sec:DSEE}

Adapting pre-trained models with $\Delta\mathcal{W}_l$ and $\Delta\mathcal{W}_s$ can bring significant parameter-efficiency, but does not directly bring any resource-efficiency such as memory or computational cost. Motivated by such, we propose a unified framework called DSEE pursuing both parameter- and resource-efficiency simultaneously. We leverage the sparsity in pre-trained models' weights to enhance the resource efficiency, as demonstrated in Figure~\ref{fig:methods}. More specifically, we derive sparse masks $\mathcal{M}$ \textit{directly} from the parameter-efficient updates $\Delta\mathcal{W}$, and apply the sparse masks by pruning the pre-trained weights $\mathcal{W}$ to seek resource-efficiency. It requires no additional parameter for sparsifying the model and no access to the underlying pretrained weights, which is favorable due to the lower sparsification cost. 

As shown in Algorithm~\ref{alg:DSEE}, DSEE handles unstructured and structured pruning at the same time: for \underline{unstructured pruning}, we sort the magnitude of $\Delta\mathcal{W}$, generate a sparse mask $\mathcal{M}$ by assigning ``1'' to the position where $\Delta\mathcal{W}$ have largest magnitude and ``0'' to the rest; for \underline{structured pruning}, we sum the magnitude of $\Delta\mathcal{W}$ of each head and remove those with least scores. We also shrink $\Delta\mathcal{W}$ accordingly by removing the corresponding weight columns in $\mathcal{V}$ and $\Delta\mathcal{W}_s$ to match the shape while keeping $\mathcal{U}$ intact. A comparison of different pruning criteria is shown in Section~\ref{sec:score_function}, which demonstrates that $\Delta\mathcal{W}$ is a superior choice due to the high downstream task performance and no access to the pretrained weights $\mathcal{W}$.

Given a parameter budget, the number of parameters per module decreases if we choose to adapt more modules, which imposes a trade-off. 
We study different choices of modules to adapt in Section~\ref{sec:adapted_modules}, and we find the optimal modules to adapt are $W_q$ and $W_v$, where $W_q$ and $W_v$ stand for the projection weights for query and value in attention heads. Since some modules are not adapted during fine-tuning (\textit{i.e.}, $\Delta\mathcal{W}=0$), we prune them separately according to the magnitude of the corresponding pre-trained weights. 
After applying the mask $\mathcal{M}$ to the pretrained weights $\mathcal{W}$, we conventionally tune $\Delta\mathcal{W}_l(=\mathcal{U}\mathcal{V})$ and $\Delta\mathcal{W}_s(=\mathcal{P}_\Omega(\mathcal{S}))$ for several epochs to recover the performance~\cite{han2015deep}. 

\begin{algorithm} 
\caption{DSEE}
\SetNoFillComment
\label{alg:DSEE}
\KwIn{Pretrained weights $\mathcal{W}$, number of non-zero elements $N$, desired sparsity $s$, loss function $\mathcal{L}$. }
\KwOut{Sparse mask $\mathcal{M}$, matrices $\mathcal{U},\mathcal{V},\mathcal{S}$. }
\begin{algorithmic}
\STATE Derive $\Omega$ from pretrained weights $\mathcal{W}$. 
\STATE Initialization: $\mathcal{U}=0,\mathcal{V}\sim\mathcal{N}(0,0.02)$, and $\mathcal{S} = 0.$ 

\tcc{{\small I: train before pruning}}
\STATE Train $\mathcal{U}, \mathcal{V}, \mathcal{S}$ with respect to $\mathcal{L}$ under the constraint of $P_{\Omega^C}(\mathcal{S})=0$.  

\tcc{{\small II: pruning the model}}
\IF{using unstructured pruning}
\STATE Prune $(1-s\%)$ weights in $\mathcal{W}$ by \quad \quad 
 sorting the magnitude of $\Delta\mathcal{W}$.
\ELSE
\STATE Prune $(1-s\%)$ heads by sorting the aggregated magnitude of $\Delta\mathcal{W}$ of heads.
\STATE Shrink $\mathcal{V}$ and $\mathcal{S}$ accordingly to match the shape. 
\ENDIF

\tcc{{\small III: tuning after pruning}}
\STATE Fine-tune $\mathcal{U}, \mathcal{V}, \mathcal{S}$ to recover the performance. 
\end{algorithmic}

\end{algorithm}







\vspace{-2mm}
\section{Experiment Results}
\vspace{-2mm}
\label{sec:prelim}
\paragraph{Datasets and models.}
We use three classical pre-trained language models in our experiments: BERT$_\mathrm{BASE}$~\citep{devlin2019bert}, RoBERTa$_\mathrm{LARGE}$~\citep{liu2019roberta} and GPT-2~\citep{radford2019language}, which have $12/24/24$ layers with hidden size of $768/1024/1024$ and $110/380/354$M trainable parameters, respectively. For BERT and RoBERTa, we evaluate on the GLUE benchmarks~\citep{wang2018glue}, and for GPT-2 we use E2E~\citep{novikova2017e2e}, WebNLG~\citep{gardent2017webnlg} and DART~\citep{nan2021dart}. 
\vspace{-2mm}
\paragraph{Training and evaluation details.}  For BERT and RoBERTa, we follow the default settings in \citet{wolf2019huggingface,devlin2019bert}. We use the AdamW~\citep{loshchilov2017decoupled} optimizer for downstream fine-tuning, and a batch size of 32 for BERT and RoBERTa, and a batch size of 2 for GPT-2. 
The rest hyper-parameters for training are reported in Table~\ref{tab:setting}. 

\vspace{-2mm}
\paragraph{Evaluation Metrics.} For the GLUE benchmark, we report the accuracy, Matthew's correlation, and Pearson's correlation in the evaluation. On GPT-2, we use BLEU~\citep{papineni2002bleu}, METEOR~\citep{denkowski2014meteor}, TER~\citep{snover2006study} and NIST~\citep{doddington2002automatic} as the evaluation metrics. To evaluate the efficiency of models, we report the number of trainable parameters to measure the parameter-efficiency, the number of total parameters (the number of non-zero parameters in the model) to measure the resource-efficiency, and FLOPs for the computational efficiency. 

\vspace{-2mm}
\paragraph{Baselines.} 
On BERT and RoBERTa, we conduct comprehensive experiments with the following baseline methods: \ding{182} Fine-tune: directly fine-tuning the full model; \ding{183} EarlyBERT~\citep{chen2021earlybert}: learn importance scores for heads and perform pruning based on them afterwards; \ding{184} BERT Tickets~\citep{chen2020lottery}: IMP-based unstructured pruning; \ding{185} P-Tuning v2~\cite{liu2021p}; \ding{186} Bitfit~\citep{zaken2021bitfit}: fine-tuning bias terms only; and \ding{187} LoRA: low-rank decomposition, which learns $\Delta\mathcal{W}_l$ only~\citep{hu2021lora}. On GPT-2, we conduct comparisons with multiple baseline methods: \ding{182} Adapters~\citep{houlsby2019parameter}: insert adapters after linear layers; \ding{183} FT-Top2: fine-tune the top 2 layers only; \ding{184}: Prefix: prefix tuning introduced by~\citet{li2021prefix}; and \ding{185} LoRA. 

\vspace{-2mm}
\subsection{Efficient Tuning with DSEE}
\vspace{-2mm}
\paragraph{Parameter-efficiency with sparse residuals. } 

To verify that using a simple low-rank component $\Delta\mathcal{W}_l$ has limitations, we compare its performance with our sparsity-embedded efficient fine-tuning. Table~\ref{tab:performance_bert_uv} shows that on four benchmarks (\textit{i.e.}, SST-2, RTE, CoLA, and MRPC), adding a sparse residual in weight updates can bring performance gain: at the level of approximately $600$K trainable parameters, adding sparse residuals with only 384 nonzero elements ($12\times2\times16=384$) can increase the validation performance on all benchmarks except CoLA by $0.23\%\sim1.09\%$; at the level of approximately $300$K trainable parameters, adding sparse residuals can bring performance gain ranged from $0.34\%$ to $1.08\%$ on all four benchmarks. 

\begin{table}
    \centering·
    \caption{Performance comparison with BERT$_\mathrm{BASE}$ on SST-2, RTE, CoLA, and MRPC. We report both the median and the standard deviation from five runs. }
    \resizebox{0.46\textwidth}{!}{
    \begin{tabular}{l|c|cccc}
        \toprule
        \multirow{2}{*}{$\Delta\mathcal{W}=$} & \# Trainable & \multirow{2}{*}{SST-2} & \multirow{2}{*}{RTE} & \multirow{2}{*}{CoLA} & \multirow{2}{*}{MRPC} \\ 
        & Parameters & & & & \\
        \midrule 
        $\Delta\mathcal{W}_l$ & 589.8K & 92.55 (0.35) & 68.95 (2.02) & 60.34 (1.69) & 86.27 (0.88)  \\
        $\Delta\mathcal{W}_l + \Delta\mathcal{W}_s$ & 590.2K & \textbf{92.78 (0.34)} & \textbf{70.04 (1.35)} & 60.31 (1.04) & 86.52 (0.57) \\
        $\Delta\mathcal{W}_l$ & 294.9K & 92.32 (0.36) & 68.23 (1.43) & 58.48 (1.61) & 86.52 (0.72) \\ 
        $\Delta\mathcal{W}_l + \Delta\mathcal{W}_s$ & 295.3K & \textbf{92.66 (0.06)} & \textbf{69.31 (2.08)} & \textbf{58.85 (0.92)} & \textbf{87.01 (0.79)} \\
        
        \bottomrule
    \end{tabular}}
    \vspace{-5mm}
    \label{tab:performance_bert_uv}
\end{table}

We further verify that adding sparse residuals $\Delta \mathcal{W}_s$ could benefit NLG tasks with GPT-2. Table~\ref{tab:performance_gpt2_uv} shows that under different levels of parameters, adding sparse residuals $\Delta\mathcal{W}_s$ yields higher performance for most metrics on three tasks. At the level of $0.39$M parameters, adding sparse residuals can improve all metrics on WebNLG and DART, and sightly boost the NIST score on E2E. At the level of $0.20$M parameters, $\Delta\mathcal{W}_s$ helps increase all metrics across three tasks. We also show the standard deviation in Table~\ref{tab:variance}. 


\begin{table*}[htbp]
    \centering
    \caption{Performance comparison of different decomposition on GPT-2 with different weight update terms. We report the median value of BLEU, MET, NIST and TER from five runs. } 
    \resizebox{0.9\textwidth}{!}{
    \begin{tabular}{l|c|ccc|ccc|ccc}
        \toprule
        \multirow{2}{*}{Forms} & \# Trainable & & E2E & & & WebNLG & & & DART & \\
        & Parameters & BLEU & MET & NIST & BLEU & MET & TER & BLEU & MET & TER  \\
        \midrule
        $\Delta\mathcal{W}=\Delta\mathcal{W}_l$  & 0.39M  & 70.38 & 46.89 & 8.844 & 55.29 & 0.414 & 0.394 & 48.23 & 0.392 & 0.469 \\ 
        $\Delta\mathcal{W}=\Delta\mathcal{W}_l+\Delta\mathcal{W}_s$ & 0.39M & 70.29 & 46.65 & 8.858 & 55.50 & 0.416 & 0.392 & 48.17 & 0.397 & 0.467\\
        \midrule
        $\Delta\mathcal{W}=\Delta\mathcal{W}_l$  & 0.20M  & 69.17 & 45.90 & 8.741 & 55.23 & 0.413 & 0.396 & 46.49 & 0.387 & 0.477 \\ 
        $\Delta\mathcal{W}=\Delta\mathcal{W}_l+\Delta\mathcal{W}_s$ & 0.20M & 69.70 & 46.85 & 8.824 & 55.56 & 0.413 & 0.392 & 47.47 & 0.393 & 0.475 \\ 
        \bottomrule
    \end{tabular}}%
    \label{tab:performance_gpt2_uv}
\end{table*}
\vspace{-1mm}
\paragraph{Resource- and parameter-efficiency with unstructured sparse masks.}

We verify that DSEE is capable of enhancing both parameter- and resource-efficiency, while preserving performance on downstream tasks, on various architectures. Table~\ref{tab:performance_bert} summarizes the experiment results on BERT$_\mathrm{BASE}$, and we observe that 
introducing unstructured sparsity patterns inside pretrained weights not only brings resource-efficiency (manifested by the fewer number of total parameters) but also potentially improves the performance on downstream tasks. Specifically, at $80\%$ and $70\%$ of total parameters, DSEE can remain comparable performance on downstream tasks, and even present a performance boost on QQP, RTE, and SST-2 compared to LoRA. At the level of $50\%$ parameters, performance on smaller datasets such as CoLA and RTE drops by a wider margin; but on larger datasets such as QQP, DSEE can maintain comparable performance ($<1.5\%$ gap) after sparsification.

\begin{table*}[ht]
    \centering
    \caption{Performance comparison of different methods on GLUE benchmarks with BERT$_\mathrm{BASE}$. We use the unstructured pruning and report the median value from five runs. $\dag$: results taken from ~\citet{chen2020lottery}. }
    \resizebox{1\textwidth}{!}{
    \begin{tabular}{l|cc|ccccccccccc}
        \toprule
        \multirow{2}{*}{Methods} & \# Trainable & \# Total & \multicolumn{8}{c}{Dataset} \\
        & Parameters & Parameters & CoLA & STS-B & MNLI & QQP & QNLI & MRPC & RTE & SST-2 \\
        \midrule
        Fine-tune$^\dag$ & $110$M & 100\% & 54.5 & 88.4 & 82.4 & 90.2 & 89.1 & 85.2 & 66.2 & 92.1 \\
        BERT Tickets$^\dag$ & $33\sim55$M & 
        $30\sim50\%$ & 53.8 & 88.2 & 82.6 & 90.0 & 88.9 & 84.9 & 66.0 & 91.9\\ 
        P-Tuning v2 & $0.3$M & $100\%$ & 59.37 & 89.36 & 82.15 & 88.50 & 90.59 & 84.80 & 67.51 & 92.20 \\
        Bitfit & $0.1$M & $100\%$ & 58.61 & 88.74 & 78.80 & 85.93 & 89.22 & 87.55 & 72.20 & 92.07\\
        LoRA & $0.6$M & $100\%$ & 59.99 & 89.09 & 83.32 & 89.48 & 90.72 & 86.27 & 68.95 & 92.32 \\ 
        \midrule
        DSEE & $0.6$M & $80\%$ & 59.94 & 89.22 & 83.29 & 90.00 & 90.46 & 86.27 & 70.76 & 92.66 \\
        DSEE & $0.6$M & $70\%$ & 58.69 & 89.08 & 83.09 & 89.97 & 90.68 & 86.27 & 71.48 & 91.97 \\
        DSEE & $0.6$M & $50\%$ & 48.49 & 87.72 & 81.84 & 89.55 & 90.12 & 81.13 & 63.90 & 91.17 \\
        \bottomrule
    \end{tabular}}%
    \vspace{-2mm}
    \label{tab:performance_bert}
\end{table*}

On GPT-2, we observe a similar trend as shown in Table~\ref{tab:performance_gpt2}. DSEE can achieve superior performance with unstructured sparse patterns with $80\%$ total parameters compared to finetuning the entire model, and remain highly competitive with other baselines with fewer parameters in the model. Using only $50\%$ of parameters in pre-trained weights, DSEE can achieve comparable performance with the full fine-tuning on E2E and DART. 



\begin{table*}[ht]
    \centering
    \caption{Performance comparison of different methods on GPT-2 on E2E, WebNLG and DART. $\ddag$: Results taken from \citet{hu2021lora}.  }
    \resizebox{0.93\textwidth}{!}{
    \begin{tabular}{l|cc|ccc|ccc|ccc}
        \toprule
        \multirow{2}{*}{Methods} & \# Trainable & \# Total & & E2E & & & WebNLG & & &  DART & \\
        & Parameters & Parameters & BLEU & MET & NIST & BLEU & MET & TER & BLEU & MET & TER\\
        \midrule 
        Fine-tune$^\ddag$ & 354.92M & $100\%$ & 68.2 & 0.462 & 8.62 & 47.6 & 0.39 & 0.50 &  46.0 & 0.39 & 0.46 \\
        Adapters$^\ddag$ & 11.48M & $100\%$ & 68.9 & 0.461 & 8.71 & 55.2 & 0.41 & 0.39 & 45.4 & 0.38 & 0.46 \\
        FT-Top2$^\ddag$ & 25.19M & $100\%$ & 68.1 & 0.460 & 8.59 & 33.5 & 0.26 & 0.75 & 38.1 & 0.34 & 0.56 \\
        Prefix$^\ddag$ & 0.35M & $100\%$ & 69.7 & 0.461 & 8.81 & 54.4 & 0.41 & 0.41 & 45.7 & 0.38 & 0.46 \\
        LoRA$^\ddag$ & 0.39M & $100\%$ & 70.4 & 0.468 & 8.85 & 55.3 & 0.41 & 0.39 & 47.5 & 0.39 & 0.45 \\
        \midrule
        
        DSEE & 0.39M & $80\%$ & 69.4 & 0.465 & 8.78 & 54.9 & 0.44 & 0.39 & 47.5 & 0.39 & 0.46  \\ 
        DSEE & 0.39M & $50\%$ & 69.5 & 0.466 & 8.74 &  42.0 & 0.33 & 0.53 & 43.4 & 0.37 & 0.51 \\ 
        \bottomrule    \end{tabular}}%
    \label{tab:performance_gpt2}
\end{table*}

Finally, we validate if DSEE can work on the larger model RoBERTa$_\mathrm{LARGE}$. We conduct experiments on four datasets (CoLA, SST-2, QNLI, and RTE), and present the results in Table~\ref{tab:performance_roberta}. 
Compared to full-finetuning, LoRA, and Adapter, our method reaches comparable performance on these four downstream tasks and saves resources at the same time. The performance gap is maximal $1\%$ but $30\%$ parameters in the models are removed. 


\begin{table}[ht]
    \centering
    \caption{Performance comparison of different methods on RoBERTa$_\mathrm{LARGE}$ on CoLA, SST-2, MRPC and RTE. $\ddag$: Results taken from \citet{hu2021lora}.}
    \resizebox{0.48\textwidth}{!}{
    \begin{tabular}{l|cc|cccc}
        \toprule
        \multirow{2}{*}{Methods} & \# Trainable & \# Total in & \multicolumn{4}{c}{Dataset} \\
        & Parameters & Parameters & CoLA & SST-2 & QNLI & RTE  \\
        \midrule
        Fine-tune$^\ddag$ & $355.0$M & $100\%$ & 68.0 & 95.1 & 94.7 & 86.6 \\        
        Adapter$^\ddag$ & 0.8M & $100\%$ & 66.3 & 96.3 & 94.7 & 72.9\\
        LoRA$^\ddag$ & 0.8M & $100\%$ & 68.2 & 96.2 & 94.8 & 85.2 \\ 
        \midrule
        DSEE & 0.8M & 70\% & 67.2 & 96.1 & 94.4 & 84.9 \\
        
        \bottomrule
    \end{tabular}}
    \label{tab:performance_roberta}
\end{table} 
 
\vspace{-1mm}
\paragraph{Resource- and parameter-efficiency with structured sparse masks.}

DSEE can directly perform structured pruning on weights without additional parameters such as importance scores of heads. In Table~\ref{tab:performance_bert_structure} we show the performance of structured pruned BERT$_\mathrm{BASE}$ on several tasks in the GLUE benchmark, where we study the testing accuracy after removing $3$, $6$ and $9$ attention heads on SST-2, MNLI, QNLI and QQP, as well as the inference FLOPs ratios of the model. Firstly, removing $3$ heads from the model reaches comparable performance against full fine-tuning (improved on SST-2, MNLI, and QNLI) and LoRA (improved on SST-2 and QQP), while taking advantage of reduced inference FLOPs. Secondly, removing $6$ heads from the model will lead to lower performance since half of the parameters in the projection matrices are eliminated. However, the performance of DSEE is still higher than EarlyBERT. Lastly, DSEE with $9$ heads removed from the model leads to comparable performance with EarlyBERT, but the number of trainable parameters is substantially smaller ($0.6$M versus $66$M).

\begin{table}[ht]
    \centering
    \caption{Performance comparison of different methods on GLUE benchmarks with BERT$_\mathrm{BASE}$. We perform the structured pruning and report the median value from five runs.  $\dag$: results taken from ~\citet{chen2020lottery}. }
    \resizebox{0.48\textwidth}{!}{
    \begin{tabular}{l|cc|cccc}
        \toprule
        Methods & FLOPs & \# Trainable & SST-2 & MNLI & QNLI & QQP \\
        \midrule
        Fine-tune$^\dag$ & $1.0\times$ & 110M & 92.1 & 82.4 & 89.1 & 90.2 \\
        LoRA & $1.01\times$ & 0.6M & 92.32 & 83.32 & 90.72 & 89.48 \\ 
        EarlyBERT & $0.63\times$ & $\sim 66$M& 90.71 & 81.81 & 89.18 & 90.06 \\
        DSEE (3 heads) & $0.92\times$ & 0.6M & 92.55 & 83.25 & 90.65 & 89.84 \\
        DSEE (6 heads) & $0.84\times$ & 0.6M & 92.32 & 82.32 & 90.01 & 89.11\\
        DSEE (9 heads) & $0.75\times$ & 0.6M & 91.63 & 80.02 & 88.39 & 88.56 \\
        \bottomrule
    \end{tabular}}%
    \label{tab:performance_bert_structure}
\end{table}
\vspace{-2mm}
\subsection{Ablation and Visualization}
\vspace{-1mm}
\label{sec:abl}
We study several choices of parameters and provide visualization in this section. 

\subsubsection{Different criteria for sparse masks}

\label{sec:score_function}

We find the magnitude of weight updates (\textit{i.e.}, $|\Delta\mathcal{W}|$) is an effective solution for preserving performance with both unstructured and structured pruning. 
We conduct experiments on the adapted weights (\textit{i.e.}, $W_q$ and $W_v$), and compare against two baselines: \ding{182} Random: perform random pruning on the adapted modules; \ding{183} $|\mathcal{W} + \Delta\mathcal{W}|$: perform pruning based on the magnitude of final adapted weights. 
Table~\ref{tab:scoring} shows the results on RTE and SST-2 with BERT$_\mathrm{BASE}$. We can see from the table that: \ding{182} performing unstructured pruning without accessing the pretrained weights can achieve comparable performance on RTE and SST-2, only slightly weaker than pruning with final adapted weights; \ding{183} performing structured pruning according to $\Delta\mathcal{W}$ yields the highest performance on both datasets after training. These observations verify the effectiveness of our proposal. 



\begin{table}[ht]
    \centering
    \caption{Performance of using different pruning criteria to generate unstructured masks. We only perform pruning on $W_q$ and $W_v$. The first part applies unstructured pruning and the latter applies structured pruning. }
    \label{tab:scoring}
    \begin{tabular}{c|c|c}
    \toprule
        Criterion & RTE & SST-2 \\
        \midrule
        $|\Delta\mathcal{W}|$ & 69.68 (1.37) & 91.97 (0.26)\\ 
        $|\mathcal{W} + \Delta\mathcal{W}|$ & \textbf{70.76 (2.09)} & \textbf{92.78 (0.39)}\\
        Random & 64.62 (2.28) & 91.63 (0.25)\\ 
        \midrule \midrule $|\Delta\mathcal{W}|$ & \textbf{70.40 (1.05)} & \textbf{92.55 (0.43)}\\
        $|\mathcal{W} + \Delta\mathcal{W}|$ & 68.59 (1.60) & 92.20 (0.60) \\
        Random & 68.23 (1.29) & 91.97 (0.14) \\
        \bottomrule
    \end{tabular}
    
\end{table}
\vspace{-2mm}
\subsubsection{Different choices of modules to adapt}
\vspace{-1mm}
\label{sec:adapted_modules}
We study the choices of modules to adapt for DSEE on RTE. We choose possible modules to adapt within $W_q$, $W_k$, $W_v$, and $W_o$, representing the projection matrix for query, key, value, and output, respectively. We hold the number of trainable parameters at the same level and set the sparsity level at $30\%$. Table~\ref{tab:modules_to_adapt} summarizes the performance with different adapted weights, which demonstrates that adapting $W_q$ and $W_v$ yields the highest performance. Each module will be given fewer parameters when adapting more modules and the model may not be sufficiently fine-tuned when adapting fewer modules and leading to inferior performance.  

\begin{figure}[ht!]
    \centering
    \includegraphics[width=0.47\textwidth]{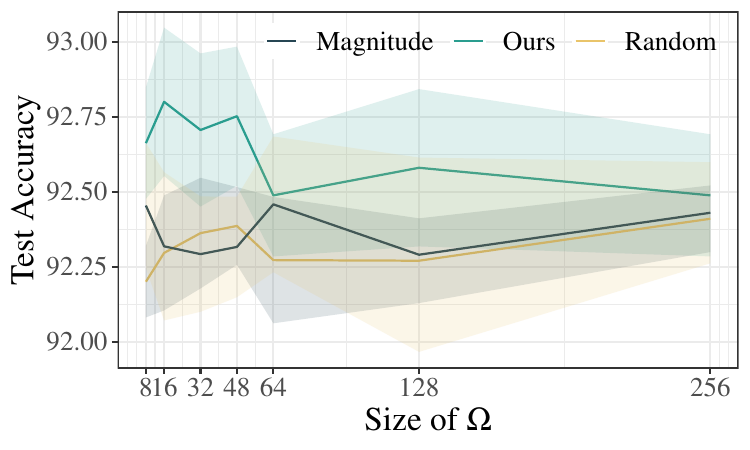}
    \caption{Testing performance on SST-2 with different sizes of $\Omega$. We report the average accuracy and the $90\%$ confidence interval of five runs. }
    \label{fig:mask_ablation}
\end{figure}

\textbf{Different methods for identifying $\Omega$.} We compare our proposal against various methods to identify $\Omega$ from pretrained weights $\mathcal{W}$: \ding{182} \textit{Magnitude}, which selects the position of elements with highest magnitude into $\Omega$; \ding{183} \textit{Random}, which randomly samples positions into $\Omega$. The results are shown in Figure~\ref{fig:mask_ablation}. We can observe that our proposal can identify high-quality $\Omega$ for finetuning on downstream tasks, shown by the consistently higher performance with different sizes of the indices set $\Omega$.

\textbf{Different sizes of $\Omega$.} We search over $8\sim256$ to find the optimal size of $\Omega$. $\Omega$ with a smaller size brings fewer performance gains, and $\Omega$ with a larger size may harm the efficiency. Figure~\ref{fig:mask_ablation} shows the relationship between the size of $\Omega$ and the performance on SST-2. We find the optimal choice for this task is $16$ where the model achieves the highest performance. Consequently, we by default set the size of $\Omega$ to 16 for simplicity.

\vspace{-2mm}
\section{Conclusion}
\vspace{-2mm}

This paper draws on the prior of sparsity and establishes the DSEE framework. It is the first attempt toward jointly optimizing both parameter-efficiency of the fine-tuning process, and the resource-efficiency of the fine-tuned model. On state-of-the-art large-scale language models (e.g., BERT, GPT, and RoBERTa) and across several datasets, DSEE consistently demonstrates highly impressive parameter and inference efficiency, in addition to preserving a competitive downstream transfer performance on various tasks. Our future work targets extending DSEE to the finetuning of large-scale computer vision and/or multi-modal pre-trained models.

\paragraph{Limitation}
The unstructured sparse patterns we introduce are not as hardware-friendly as the structured patterns, suggesting the speedup of using unstructured patterns maybe limited due to the implementation. The number of parameters of models we are studying are only at the level of $100\sim 300$M, and the datasets are focus on GLUE, E2E, WebNLG, and DART. We will generalize to wider choices of datasets in future works. 
\section{Ethical and Broader Impacts}
DSEE aims at reducing the number of trainable parameters when fine-tuning the models, which can help save the cost of saving new weights. This can be helpful to companies who are fine-tuning large-scale language models on various downstream tasks, suggesting our work has potentially positive broader impact. On the other hand, our work does not have obvious ethical impacts, as we focusing on model tuning. 

\clearpage

\bibliography{anthology,custom}
\clearpage
\appendix

\section{More Implementation Details}

\subsection{Hyper-parameters}
\label{sec:appendix}

We report the learning rates, the batch sizes, and the max sequence length for DSEE in Table~\ref{tab:setting}. The device we used for experiments are various, including NVIDIA GeForce GTX 1080 Ti, GeForce RTX 2080 Ti, Titan RTX, and A6000. We follow~\cite{hu2021lora} to set the evaluation protocols on E2E, WebNLG, and DART.

\subsection{Decomposition Method}
\label{sec:decompose}
GreBsmo~\citep{zhou2013greedy} is an algorithm for solving the Robust PCA-like methods. The optimization of $\mathcal{U}$, $\mathcal{V}$ and $\mathcal{S}$ follows the following iterative rules:

\begin{equation}
\left\{\begin{array}{l}
\label{equ:iterateUpdate}
U_{k}=Q, {\mathrm{QR}}\left(\left(X-S_{k-1}\right) V_{k-1}^{T}\right)=Q R \\
V_{k}=Q^{T}\left(X-S_{k-1}\right) \\
S_{k}=\mathcal{S}_{\lambda}\left(X-U_{k} V_{k}\right)
\end{array}\right. ,
\end{equation}
where $X$ is the original dense matrix, $\mathrm{QR}(\cdot)$ means the QR decomposition, $S_\lambda(\cdot)$ indicates the soft-threshold function (\textit{i.e.}, $S_\lambda(x)=x \mathbf{1}_{|x|\ge\lambda}$) , and the subscripts $k$ indicates the optimization step. 
\subsection{Statistics and Usage of Datasets}

We report the statistics of datasets in Table~\ref{tab:statistics}. For GLUE tasks we report the sizes of the train, the dev and the test set, and for non-GLUE tasks we report the sizes of the train, validation (dev), and test set. We follow the conventional use of these datasets~\cite{hu2021lora} and do not modify the conventional splits.

\begin{table}[h]
    \centering
    \caption{The statistics of datasets we used for experiments. }
     \label{tab:statistics}
    \begin{tabular}{c|c|c|c}
    
    \toprule Name & Train & Dev & Test \\ \midrule
    \multicolumn{4}{c}{GLUE} \\ \midrule
    CoLA &  8,551 & 1,043 & -\\
    SST-2 & 67,349 & 872 & - \\
    MNLI & 392,702 & 9,815 & - \\
    QNLI & 104,743 & 5,463 & - \\
    QQP & 363,846 & 40,430 & - \\
    STS-B & 5,749 & 1,500 & - \\
    RTE & 2,490 & 277 & - \\
    MRPC & 3,668 & 408 & - \\
    \midrule
    \multicolumn{4}{c}{non-GLUE} \\
    \midrule
    E2E & 42,061 & 4,672 & 4,693\\
    WebNLG & 18,025 & 2,258 & 4,928 \\
    DART & 30,526 & 2,768 & 6,959 \\
    \bottomrule
    \end{tabular}
\end{table}

\section{More Experiments Results}

\subsection{Ablation Studies}
 Table~\ref{tab:modules_to_adapt} summarizes the performance with different adapted weights, which demonstrates that adapting $W_q$ and $W_v$ leads to the highest performance. 
\begin{table}[h]
    \centering
    \caption{Testing performance of BERT$_\mathrm{BASE}$ on RTE with different adapted modules. We report the median values and the standard deviation from three runs. }
    \label{tab:modules_to_adapt}
    \resizebox{!}{0.16\linewidth}{
    \begin{tabular}{c|c|c|c}
    \toprule
        Weights & Test Acc. & Weights & Test Acc. \\
        \midrule
        $W_q$ & 68.59 (0.21) & $W_k$ & 67.87 (0.21)\\
        $W_v$ & 68.23 (1.82) & $W_o$ & 68.23 (1.05) \\
        $W_q$,$W_k$ & 68.95 (1.11) & $W_q$,$W_v$ & \textbf{71.48 (2.16)} \\
        $W_k$,$W_v$ & 70.04 (0.75) & $W_q$,$W_k$,$W_v$ & 69.31 (2.56) \\
        \bottomrule
    \end{tabular}}
    \vspace{-2mm}
\end{table}

\begin{table*}[htbp]
    \centering
    \caption{ Performance comparison of different decomposition on GPT-2 with different weight update terms. We report the standard deviation of BLEU, MET, NIST and TER from five runs. } 
    \resizebox{0.9\textwidth}{!}{
    \begin{tabular}{l|c|ccc|ccc|ccc}
        \toprule
        \multirow{2}{*}{Forms} & \# Trainable & & E2E & & & WebNLG & & & DART & \\
        & Parameters & BLEU & MET & NIST & BLEU & MET & TER & BLEU & MET & TER  \\
        \midrule
        $\Delta\mathcal{W}=\Delta\mathcal{W}_l$  & 0.39M  & 0.43 & 0.13 & 0.037 & 0.37 & 0.005 & 0.003 & 0.23 & 0.001 & 0.001 \\ 
        $\Delta\mathcal{W}=\Delta\mathcal{W}_l+\Delta\mathcal{W}_s$ & 0.39M & 0.07 & 0.26 & 0.047 & 0.48 & 0.005 & 0.004 & 0.40 & 0.003 & 0.002 \\
        \midrule
        $\Delta\mathcal{W}=\Delta\mathcal{W}_l$  & 0.20M  & 0.23 & 0.03 & 0.043 & 0.26 & 0.005 & 0.007 & 0.06 & 0.002 & 0.001 \\ 
        $\Delta\mathcal{W}=\Delta\mathcal{W}_l+\Delta\mathcal{W}_s$ & 0.20M & 0.61 & 0.19 & 0.029 & 0.52 & 0.006 & 0.004 & 0.15 & 0.001 & 0.001 \\ 
        \bottomrule
    \end{tabular}}%
    \label{tab:variance}
\end{table*}

\begin{table*}[t]
    \centering
    \caption{Hyper-parameters we used on different datasets and architectures. }
    \label{tab:setting}
    \resizebox{0.96\textwidth}{!}{
    \begin{tabular}{cc|c|ccccccccc}
        \toprule
        \multirow{2}{*}{Architecture} & \multirow{2}{*}{Method} & \multirow{2}{*}{Parameters} & \multicolumn{8}{c}{Dataset} \\
        & & & MNLI & QNLI & QQP & SST-2 & CoLA & MRPC & RTE & STS-B  \\
        \midrule
        BERT$_\mathrm{BASE}$ & DSEE (before pruning) & Learning Rate & 2e-4 & 2e-4 & 2e-4 & 2e-4 & 1e-3 & 8e-4 & 6e-4 & 8e-4 \\
        BERT$_\mathrm{BASE}$ & DSEE (after pruning) & Learning Rate & 2e-4 & 2e-4 & 2e-4 & 2e-4 & 1e-3 & 8e-4 & 6e-4 & 8e-4 \\
        BERT$_\mathrm{BASE}$ & DSEE & Batch Size & \multicolumn{8}{c}{32}\\
        BERT$_\mathrm{BASE}$ & DSEE & Max Sequence Length & \multicolumn{8}{c}{128}\\
        \midrule
        RoBERTa$_\mathrm{LARGE}$ & DSEE (before pruning) & Learning Rate & - & 2e-4 & - & 4e-4 & 3e-4 & - & 4e-4 & - \\ 
        RoBERTa$_\mathrm{LARGE}$ & DSEE (after pruning) & Learning Rate & - & 2e-4 & - & 4e-4 & 3e-4 & - & 4e-4 & - \\ 
        RoBERTa$_\mathrm{LARGE}$ & DSEE & Batch Size & - & 32 & - & 32 & 16 & - & 32 & - \\ 
        RoBERTa$_\mathrm{LARGE}$ & DSEE & Max Sequence Length & - & 512 & - & 512 & 128 & - & 512 & - \\ 
        \bottomrule
    \end{tabular}}%
\end{table*}

\textbf{Performance of different $\Omega$.} We conduct additional ablation study experiments (three runs for each experiment) on the sizes of $\Omega$ on three datasets in GLUE (\textit{i.e.}, STSB, QNLI and MRPC). The results shown in Table~\ref{tab:size_omega_performance} below verify that our method are generalizable to other datasets. On STSB and QNLI, using a size of 16 can achieve the best performance, while on MRPC it can achieve a comparable test accuracy.

\begin{table}[ht]
    \centering
    \caption{Performance on the three datasets using different sizes of $\Omega$.} 
    \begin{tabular}{c|c|c|c|c}
    \toprule
       Dataset  & $\Omega=8$ & $\Omega=16$ & $\Omega=32$ & $\Omega=48$ \\
        \midrule STSB &	89.26 & 	89.28 & 89.10 & 89.16 \\
        \midrule MRPC& 85.38 & 	86.27& 86.36& 86.27\\ 
        \midrule QNLI&91.01&91.06&91.00	&90.87 \\ \bottomrule
    \end{tabular}    \label{tab:size_omega_performance}
\end{table}

\textbf{Compare with recent methods.}
We have conducted a set of experiment to compare our methods with MAM Adapter~\cite{he2021towards}. We train a RoBERTa-large with their method on {SST-2, QNLI, RTE}, by following the same hyperparameters used in the original work. The 
 results are shown in Table~\ref{tab:compare_recent_methods}.
 We observe that our method, even with sparse models, achieves same-level performance with LoRA and MAM Adapter.

\begin{table}[ht]
    \centering
    \caption{Compare with more methods. } 
    \resizebox{1\linewidth}{!}{
    \begin{tabular}{c|c|c|c|c}
    \toprule
       Method&Total Parameters&SST-2&QNLI&RTE \\
\midrule LoRA&100\%&96.2&	94.8&85.2\\
MAM Adapter&100\%&96.1&94.7&80.4\\
Ours&70\%&96.1&94.4&84.9\\ \bottomrule

    \end{tabular}} \label{tab:compare_recent_methods}
\end{table}

\textbf{Other Pruning Methods.} We apply the iterative magnitude pruning method on RTE. Specifically, we train the model for $10$ epochs, prune $10\%$ of the remaining weights, and fine-tune for $10$ epochs before the next pruning. Table~\ref{tab:imp} shows that directly applying iterative magnitude pruning does not bring performance improvements over the one-shot pruning baseline.

\begin{table}[h]
    \centering
    \caption{Applying iterative magnitude pruning (IMP) to prune models. }
    \begin{tabular}{c|c}
    \toprule
        Remaining Weights & Accuracy \\ \midrule
90\%&	70.02\% \\
81\%&	70.76\% \\
72.9\%&	63.90\% \\ 
65.6\%&	61.01\%\\ \bottomrule

    \end{tabular}
    
    \label{tab:imp}
\end{table}

\end{document}